\documentclass[wcp]{jmlr}

\usepackage{longtable}

\usepackage{booktabs}

\makeatletter
\let\Ginclude@graphics\@org@Ginclude@graphics 
\makeatother

\jmlrvolume{}
\jmlryear{2024}
\jmlrworkshop{ACML 2024}

\title[Remembering Everything Makes You Vulnerable]{“Remembering Everything Makes You Vulnerable”: A Limelight on Machine Unlearning for Personalized Healthcare Sector}

  \author{\Name{Ahan Chatterjee} \Email{Chatterjee.Ahan@campus.lmu.de}\\
   \addr Ludwig-Maximilians-University, Munich, Germany
   \AND
   \Name{Sai Anirudh Aryasomayajula} \Email{A.Aryasomayajula@campus.lmu.de}\\
  \addr Ludwig-Maximilians-University, Munich, Germany
    \AND
  \Name{Rajat Chaudhari} \Email{Rajat.Chaudhari@campus.lmu.de}\\
  \addr Ludwig-Maximilians-University, Munich, Germany
   \AND
   \Name{Subhajit Paul} \Email{subhajit.paul@tnu.in}\\
  \addr The Neotia University, Kolkata, India
    \AND
   \Name{Vishwa Mohan Singh} \Email{Vishwa.Singh@campus.lmu.de}\\
  \addr Ludwig-Maximilians-University, Munich, Germany
 }

\begin{document}

\maketitle

\begin{abstract}
As the prevalence of data-driven technologies in healthcare continues to rise, concerns regarding data privacy and security become increasingly paramount. This thesis aims to address the vulnerability of personalized healthcare models, particularly in the context of ECG monitoring, to adversarial attacks that compromise patient privacy. We propose an approach termed "Machine Unlearning" to mitigate the impact of exposed data points on machine learning models, thereby enhancing model robustness against adversarial attacks while preserving individual privacy. Specifically, we investigate the efficacy of Machine Unlearning in the context of personalized ECG monitoring, utilizing a dataset of clinical ECG recordings. Our methodology involves training a deep neural classifier on ECG data and fine-tuning the model for individual patients. We demonstrate the susceptibility of fine-tuned models to adversarial attacks, such as the Fast Gradient Sign Method (FGSM), which can exploit additional data points in personalized models. To address this vulnerability, we propose a Machine Unlearning algorithm that selectively removes sensitive data points from fine-tuned models, effectively enhancing model resilience against adversarial manipulation. Experimental results demonstrate the effectiveness of our approach in mitigating the impact of adversarial attacks while maintaining the pre-trained model accuracy.\end{abstract}
\begin{keywords}
Machine Unlearning; ResNet34; ECG; FGSM; Adversarial Attack\end{keywords}

\section{Introduction}
The influx of massive data and the break of the Big Data era have resulted in the development of complex shallow and deep learning algorithms. Most of the State-of-the-Art (SOTA) models have been trained on a huge corpus of data – be it Imagenet (Deng et. al. 2009) or others. In today’s world where it’s said that privacy is a myth, there are stringent laws to ensure that user data remains in the safe hands. The privacy laws range from the General Data Protection Regulation (GDPR) of EU Nations to the California Consumer Privacy Act (CCPA), which guarantees the user to revoke the right to store the data from datasets or databases at any given time point. Now, when the data is asked to be removed from a database or any dataset it is easier to remove that particular instance of the data. Though deleting the user data from only the database doesn’t ensure security or complete removal of the data as that data might have been used to train any machine learning model. And from machine learning model, training data can be reconstructed using variety of methods including adversarial attacks. This motivates the use of Machine Unlearning where we aim to mitigate the effect of removed data point from the machine learning models so that the model doesn’t leak out sensitive data points. In this paper, we have tried to tap in the personalized healthcare sector which is way more vulnerable than any other sector in terms of sensitivity of the data which can be leaked out. We worked on personalized ECG monitoring; it’s a collection of models which is responsible to adapt and analyze ECG waves and detect any abnormal or anomalous events (basically cardiac events such as Arrhythmia etc.). Personalized ECG models are better in terms of generalized ECG models is because it minimizes the confounding effect of individual variations by comparing the electric waves to the previously trained models which have a large corpus of individual data. Over the time as the ECG accumulates they form a large corpus of the individual data, which over time fine tunes the baseline model for the individual patient, based on disease progression and medical intervention. Though all of it sounds magical the issue lies on the safety margin of the model. Adversarial attacks like FSGM can feed the model with perturbated input to assess which data points actually belong the model and reconstruct the training data. As these are fine tuned for a particular patient so the difference of data spikes in 2 models will certainly expose the underlying cardiac condition for the patient which can be widely misused from insurance companies to medicine industry. In order to make the world a safer place we aim to design a Machine Unlearning algorithm which will unlearn the individual’s medical record from the personalized model making it safer as it won’t expose sensitive data further. Along with that the model still remains personalized and expected to work better in comparison to any generalized model. In a nutshell, we aim to train a deep neural classifier on ECG Data, and we call that our generalized model and moving to further fine tune the said model for a particular individual. We have also demonstrated that how a basic FGSM attack can expose the extra data points for the fine tuned model. In order to protect it we have implemented our machine unlearning algorithm which unlearns the fine tuned data points, preventing itself from getting attacked and still providing better accuracy than any generalized model. The paper is further structured as Section 2 contains literature survey, followed by methodology in section 3 and results in section 4.  The last section i.e. section 5 provides discussion and conclusion for the paper.

\section{Related work}
Before handling over to expensive medical imaging for heart diagnosis, the first step avenue is to analyze and test the Electrocardiogram (ECG) signals. Advent of smart wearable enabled the market to advance towards personalized ECG systems. Lacks of emergency medical personals have also contributed into the growth of automated ECG Signal Analysis. (Linhai and Liang, 2022)
	The analyses of ECG signals have evolved from being trained on shallow learning techniques to modern deep learning algorithms such as Convolution Neural Networks. Initial significant work of CNN on ECG signal have been carried out by S. Kiranyaz et al. (2015) \cite{kiranyaz2015real} where they have 1D CNN for ECG classification on the MIT-BIH Dataset. On the recent works of Zhao et al. (2020) they have decomposed the entire ECG Signal into 9 sub-signals by wavelet functions and then they have reconstructed the segments which eliminates the noises in the signal which helped them to achieve 86.46\% of F1 Score. \cite{zhao2020ecg}. The work of Jun et. al. (2018) explores the classification of ECG signals using 2D Convolution networks and they outperformed the work of Kiranyaz et. al. and achieved sensitivity of 97.3\%. \cite{jun2018ecg}. The work of Baloglu et. al. (2019) proposed a novel architecture of stacked CNN blocks to classify the ECG Signals. They have considered a 12 Lead ECG Signal Data which helped for more accurate measures for Miocardial Infarction and they have achieved 99\% score. \cite{baloglu2019classification}. 

 Now, all Deep Neural Networks are mostly susceptible to adversarial attacks, where a small perturbed data input is being passed for sth number of times and the model checks for spikes around the data points on which the algorithm has been trained on. Small perturbed input can lead to change the DNN output. Adversarial perturbations can be introduced into an ECG signal after it has been acquired from a patient and before it is analyzed by a Deep Neural Network (DNN) classifier. Such actions may allow the assailant to reap personal rewards while negatively harming the patient. Szegedy et al. \cite{szegedy2014intriguing}discovered adversarial perturbations in picture categorization in 2013. This significant research uncovered flaws in Deep Neural Networks (DNNs) when subjected to well constructed, unnoticeable alterations to input photos, resulting in inaccurate model predictions. Following this original discovery, the research community developed a number of advanced strategies for producing hostile perturbations. Among these strategies is the Fast Gradient Sign Method (FGSM), developed by Goodfellow et al. in 2014 \cite{goodfellow2015explaining}, which uses the loss gradients with respect to the input picture to create perturbations. Kurakin et al. (2016) developed the Basic Iterative technique (BIM) \cite{kurakin2017adversarial}, an extension of FGSM, which iteratively uses the gradient sign technique in tiny stages, hence increasing the attack's efficacy. The Projected Gradient Descent (PGD) approach, published by Madry et al. in 2017,\cite{madry2019deep} is regarded as one of the most powerful adversarial attack methods. It employs a multi-step form of gradient descent and operates inside a given epsilon ball around the input data. Furthermore, the Carlini and Wagner (C\&W) assaults, developed by Carlini and Wagner in 2017\cite{carlini2017evaluating}, use a more complex method to producing adversarial cases by optimizing a modified objective function that includes a term to reduce the amount of the perturbation. These strategies, which are primarily aimed at weakening the integrity of DNNs in image classification tasks, highlight the vital need of designing strong machine learning models that can withstand adversarial manipulation.

 Furthermore, recent adversarial assaults on trained models have showed the capacity to determine which instances or characteristics belonged to the training data. This needs a new method known as machine unlearning, which causes machine learning models to forget certain facts. Cao et al.\cite{cao2015towards} initially introduced the concept of machine unlearning, aiming to negate the influence of a specific data point on a trained model efficiently and precisely. This concept, which ensures that removing training data does not alter the model's distribution, was later formalized by Ginart et al. \cite{ginart2019making}, who established foundational principles for designing data forgetting algorithms. These approaches, however, were primarily effective for non-adaptive ML models like k-means clustering. To address this, Bourtoule et al. \cite{bourtoule2021machine} proposed the SISA method, a model-agnostic technique that partitions training data into separate slices for individual model training, enabling precise unlearning but at increased storage costs. Golatkar et al. \cite{shi2023deepclean} further advanced the field by introducing a technique to remove weights associated with to-be-forgotten data, eliminating the need for retraining. As the field evolved, various strategies emerged for estimating and mitigating the impact of removing training data on ML models, including influence functions, weight removal, linear replacement, and gradient updating. These methods provide approximate forgetting and mathematical guarantees for certified data removal in DNNs, marking machine unlearning as a burgeoning area of research with significant implications for model flexibility and ethical data management.

\section{Methodology}
\subsection{Deep Neural Based Electrocardiogram Classifier}
\subsubsection{\textbf{Dataset Description}}

A large collection of clinical ECG recordings, the PTB-XL ECG dataset\cite{article} is renowned for its substantial size and thorough annotations. It includes 21,837 clinical 12-lead electrocardiograms (ECGs) with a 10-second length from 18,885 individuals. One or two cardiologists' structured annotations of 71 distinct ECG statements that adhere to the SCP-ECG standard make this dataset unique. These annotations provide as a valuable resource for training and assessing automated ECG interpretation algorithms as they capture diagnostic, form, and rhythm comments. This dataset is interesting since it includes a significant percentage of healthy control samples in addition to the range of illnesses it covers. Superclasses include Normal ECG, Myocardial Infarction, ST/T Change, Conduction Disturbance, and Hypertrophy are among those into which the diagnosis is divided. There is a downsampled version of the dataset at 100Hz for convenience, and it is also accessible in WFDB format with 16-bit precision and 500Hz sampling frequency.

\subsubsection{\textbf{Data Pre-processing}}

The data is processed at a sampling rate of 500 Hz, a standard measure for capturing the intricate details within ECG signals. In the dataset, we notice some ECGs with blank leads, so we filter out those data points. For the patient data point, we split the ECG data of 5000 timestamps, into 10 parts, each consisting of 500 timestamps each. This helps us fine tune the model with more data points. Additionally, a Fast Fourier Transform (FFT) analysis is defined through a function to transform the time-domain ECG signals into the frequency domain, allowing for the identification and analysis of the constituent frequencies in the ECG signal. Finally, to ensure the quality of the ECG signals, a noise filtering procedure is applied to all channels (leads) of the ECG data. This is accomplished with a moving average filter, which smoothens the signal by averaging over a window of specified size, thereby mitigating the effect of short-term fluctuations due to noise.

The entire preprocessing pipeline underscores the importance of careful data handling to preserve the integrity of the ECG signals, while also preparing them for subsequent analysis, such as the development and training of machine learning models for automatic ECG interpretation.

\subsubsection{\textbf{Model Architecture}}

ResNet-34 belongs to the family of Residual Networks (ResNet)\cite{7780459}, which He et al. proposed in 2015 to address the issue of vanishing gradients in deep neural networks, hence enabling the training of far deeper networks than was previously possible.

A deep convolutional neural network with 34 weighted layers is called ResNet-34. A sequence of residual blocks is included, each of which consists of a shortcut link that omits one or more layers. The phrase "residual network" comes from the identity mapping these connections execute, in which their outputs are added to the stacked layer outputs (also known as the residual). In contrast to conventional designs, which experience performance degradation in deeper networks beyond a certain depth, ResNet-34 makes use of these shortcut connections to facilitate the training of deep networks, without the loss of performance due to vanishing gradients. Usually, two or three convolutional layers with batch normalization and ReLU activations are present in each residual block of ResNet-34. The architecture of ResNet-34 is structured as follows: an initial convolutional layer, a sequence of residual blocks that increase the number of filters by two at each level, average pooling, and a fully connected layer for classification come last.

The use of batch normalization, which normalizes the input layer by modifying and scaling the activations, is one of ResNet-34's primary characteristics. This considerably lowers the number of epochs required to train deep networks and aids in stabilizing the learning process.

The architecture has been widely used for a range of computer vision applications and has demonstrated remarkable performance on large-scale image recognition tasks. It has sparked more study into residual learning and network depth scaling and is the foundation for many later advances in neural network architecture.

The design ideas of ResNet-34 have influenced the larger machine-learning community and established a new benchmark for creating deep-learning models that can be used for a wide range of applications, including visual recognition.

\begin{figure}[htbp]
    \centering
    \includegraphics[width=0.5\textwidth]{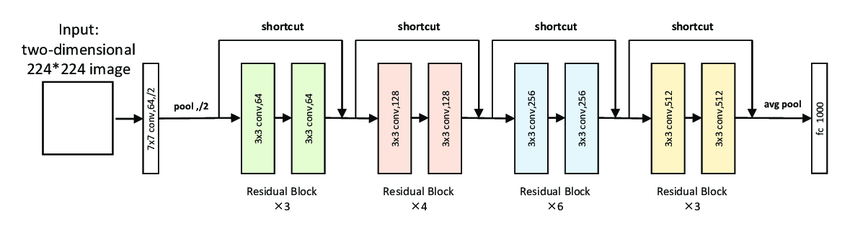}
    \caption{ResNet34 Architecture}
    \label{fig:example}
\end{figure}

\subsubsection{\textbf{Fine Tuned Model}}

The "Fine-Tune Model" portion describes how the whole ResNet-34 architecture was used for the model's first training after it was subjected to a generalized dataset. During this basic training stage, the model was exposed to a large and varied set of electrocardiogram (ECG) patterns, covering a broad spectrum of patient information. As a result, the model established a foundation for specialized learning by developing a comprehensive grasp of ECG signals.

After going through this stage, the model was adjusted using information from specific patients. This fine-tuning is a targeted type of transfer learning in which a smaller, more focused dataset related to a particular patient is used to further train the pre-trained network, which already possesses a generalized understanding of ECG readings. A few changes are usually made during the fine-tuning phase, such as decreasing the learning rate to avoid overwriting previously learned features and selectively retraining some network layers while freezing others. Using the model's pre-trained skills, this approach modifies its emphasis to more accurately identify and anticipate the distinct patterns seen in each patient's ECG data.
  \subsection{Adversarial Attack on Fine Tuned Models for Data Reconstruction}
  An essential diagnostic technique for identifying and classifying serious electrical anomalies such as myocardial infarctions and cardiac arrhythmias is electrocardiography (ECG). Many machine learning (ML) and artificial intelligence (AI) techniques have been developed to accurately recognize different ECG patterns. Although these technologies provide many benefits, state-of-the-art arrhythmia prediction systems are vulnerable to adversarial assaults. These flaws are critical because they might result in improper hospital admissions, misdiagnoses, patient data privacy violations, insurance fraud, and unfavorable consequences for medical facilities. In this section, we are going to explore Fast Gradient Sign Method (FGSM) algorithm to spike and regenerate the training data and explore how can we extract data points from the fine-tuned personalized models to identify patients on which the model has been trained. 

  \subsubsection{\textbf{Fast Gradient Sign Method (FGSM)}}
This is an example of a white box attack where we aim to feed perturbed data points to the model and assess the new data points. We're training a classifier model based on CNN data. with 12 leads and a 500 Hz frequency rate. We aim to extract data points from the personalized fine-tuned model as a part of the adversarial attack.

$E_{\text{adv}} \text{ is obtained by perturbing the original signal, } E.$

We denote the classifier defined by the CNN with softmax output activation as $\hat{y} = f(\theta, E)$ for a given input-label pair $(E, y)$. FGSM finds the adversarial image $E_{\text{adv}}$ by maximizing the loss $L(E_{\text{adv}}, y) = L(f(\theta, E_{\text{adv}}), y)$ subject to the $l_{\infty}$ perturbation constraint $\|E_{\text{adv}} - E\|_{\infty} \leq \epsilon$ with $\epsilon$ be the attack strength. We have the following approximation i.e., $L(E_{\text{adv}},y) \approx L(E,y) + \nabla_E L(E, y)^T . (E_{\text{adv}} - E)$.

\begin{equation}
    E_{\text{adv}} = E + \epsilon \cdot \text{sign} (\nabla_E L(E, y)).
\end{equation}

FGSM iterates FGSM to generate enhanced attacks, i.e.,

\begin{equation}
    E^{(m)} = E^{(m-1)} + \epsilon \cdot \text{sign} (\nabla_E L(E^{(m-1)}, y)),
\end{equation}

where $m = 1, \ldots, M$, $E^{(0)} = E$ and $E_{\text{adv}} = E^{(M)}$, $M$ being the number of iterations.

In practice, we apply the following clipped IFGSM:

\begin{equation}
    E^{(m)} = \text{Clip}_{E,\alpha} \{ E^{(m-1)} + \epsilon \cdot \text{sign} (\nabla_E L(E^{(m-1)}, y)) \},
\end{equation}

where $\alpha$ is an additional parameter to be specified in the experiments.\cite{goodfellow2015explaining}

In this detailed study, we rigorously evaluate the resilience of customized, optimized deep learning models compared to their generic counterparts under complex adversarial scenarios. Our methodology makes use of the Fast Gradient Sign Method (FGSM), a well-known adversarial attack technique that is effective at revealing neural network flaws. Our study's generalized model is painstakingly constructed by training on a large dataset that combines various patient data. By way of comparison, the optimized model gains even more refinement from an extra layer of optimization that is specifically designed to incorporate and capture the distinct features present in each patient's data.

The key to our work is the intentional introduction of well-constructed perturbations into the data that are supplied to both model types. With the use of this calculated maneuver, the discrepancy in model reactions will be triggered and then quantified, offering a precise measure of relative resilience. We have discovered a compelling pattern through rigorous empirical analysis supported by a strong theoretical foundation: the fine-tuned model consistently demonstrates a remarkable resilience, maintaining a high degree of accuracy in its predictive outputs, while the generalized model falters and gives in to the deceptive perturbations, resulting in inaccurate outputs. The points where the point of difference is occurring for perturbed data are the data points where the fine-tuned model has extra data points.

\section*{Mathematical Formulation of Proposed Algorithm Based on FGSM Attack}

Let $D_{\text{gen}}$ be the dataset encompassing a wide patient base for training the generalized model $M_{\text{gen}}$. Let $D_{\text{ft}}$ be the dataset for the fine-tuned model $M_{\text{ft}}$, which includes patient-specific data points:

\begin{equation}
    M_{\text{gen}} \gets \text{Train}(D_{\text{gen}}), \quad M_{\text{ft}} \gets \text{Train}(D_{\text{gen}} \cup D_{\text{ft}}).
\end{equation}

Given a sample $(x, y)$, we introduce an adversarial perturbation $\delta$ constrained by $\|\delta\|_{\infty} \leq \epsilon$, generating adversarial examples $x_{\text{adv}} = x + \delta$. The models' predictions can be represented as:

\begin{equation}
    \hat{y}_{\text{gen}} = M_{\text{gen}}(x_{\text{adv}}), \quad \hat{y}_{\text{ft}} = M_{\text{ft}}(x_{\text{adv}}).
\end{equation}

We define the accuracy of each model under adversarial attack as:

\begin{equation}
    \text{Acc}_{\text{gen}} = \mathbb{P}(\hat{y}_{\text{gen}} = y), \quad \text{Acc}_{\text{ft}} = \mathbb{P}(\hat{y}_{\text{ft}} = y).
\end{equation}

Our empirical analysis involves measuring the variance in the accuracy of the predictions:

\begin{equation}
    \Delta\text{Acc} = \text{Acc}_{\text{ft}} - \text{Acc}_{\text{gen}}.
\end{equation}

The discrepancy in model performance due to the additional data points in $D_{\text{ft}}$ is then:

\begin{equation}
    \text{Discrepancy} = \begin{cases}
    0, & \text{if } \Delta\text{Acc} = 0, \\
    >0, & \text{if } \Delta\text{Acc} > 0, \\
    \end{cases}
\end{equation}

The fine-tuned model $M_{\text{ft}}$ is expected to have higher accuracy for certain inputs due to its specialization, as indicated by a positive $\Delta\text{Acc}$. For the points where $\Delta\text{Acc}$ is positive are the extracted data points. Our target is to minimize the  $\Delta\text{Acc}$ through unlearning, preferably zero.

In Figure 1, two sets of ECG data are shown along with the frequency spectra that correspond to them. The original ECG signal and spectrum are displayed in the top graphs, while the perturbed ECG signal and spectrum are displayed in the bottom graphs. With discernible P,Q,R,S,and T waves indicative of a regular heartbeat, the first ECG tracing seems to be a normal signal. The majority of its energy is focused at lower frequencies in its frequency spectrum, which is characteristic of ECG signals. The waveform of the disturbed ECG signal has changed, most likely as a result of noise that was added during an adversarial assault. These modifications are also seen in the spectrum of the perturbed ECG, where the disturbance may have caused shifts in certain frequency components.

A single ECG waveform is shown in Figure 2, which has a label showing that the initial label was "Abnormal," but the model prediction following analysis is "Normal." This points to a mismatch between the expected and projected ECG signal classifications, that is the result of an adversarial attack that successfully misclassified the model or caused the model to interpret the ECG data erroneously.

\begin{figure}[htbp]
    \centering
    \includegraphics[width=0.5\textwidth]{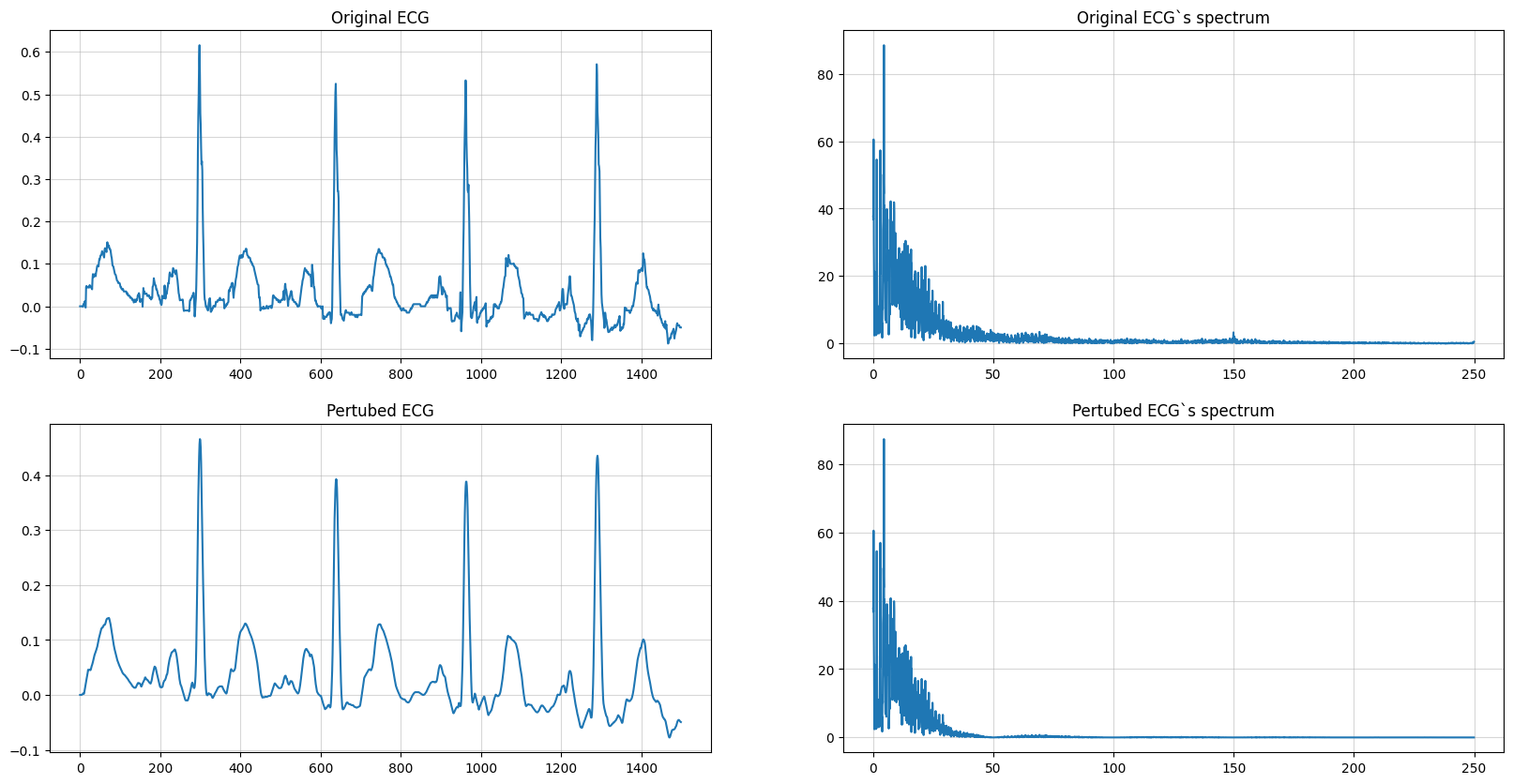}
    \caption{Original ECG and Perturbed ECG Waves}
    \label{fig: Original ECG and Perturbed ECG Waves}
\end{figure}

\begin{figure}[htbp]
    \centering
    \includegraphics[width=0.5\textwidth]{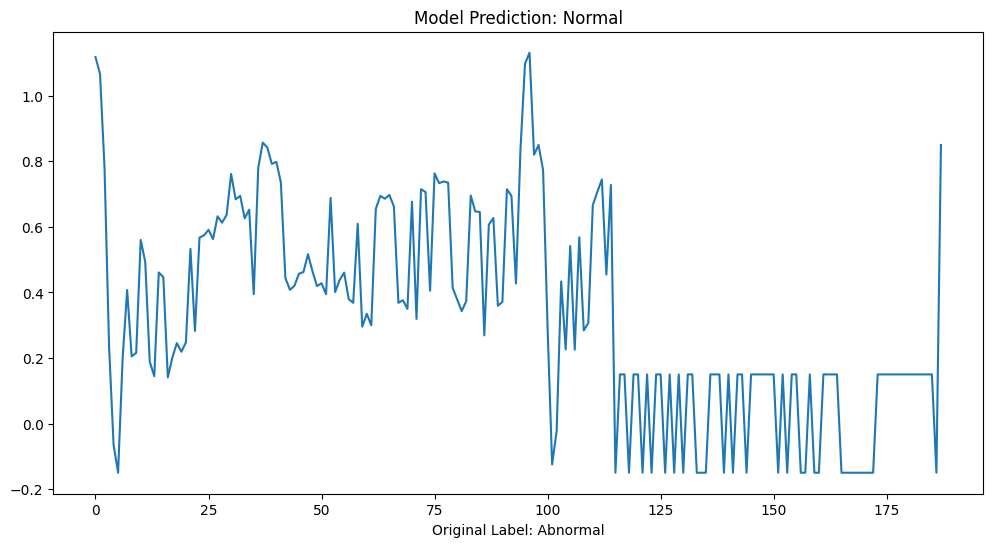}
    \caption{Model Prediction Vs. Original Label when Perturbed Data has been Injected}
    \label{fig: Pertubed ECG and Perturbed ECG Waves}
\end{figure}

\subsection{Machine Unlearning for Fine-Tuned Models}

\subsubsection{\textbf{Learning}}
Learning is the iterative process of constructing an algorithm \(A: Z^n \xrightarrow{} R\) that takes a dataset \(D\) and produces a hypothesis \(A(D)\in R\). The efficacy of \(A\) is gauged by the discrepancy between the population risk of the hypothesis \(A(D)\) and the risk of the optimal hypothesis \(r^*\) in \(R\), termed the excess risk:
\begin{center}
    \(E[F(A(D))] - F^*\)
\end{center}
where \(F\) represents the loss function and \(F^*\) denotes the minimum achievable loss attained by the minimizer \(r^*\). \newline

\subsubsection{\textbf{Machine Unlearning}}
Unlearning, as the name suggests, aims to partially reverse the learning process. Initially, we introduce the concept of a \textit{forget set} \(S \subseteq D\), denoting the collection of training examples designated for the model trained on dataset \(D\) to "forget". Conceptually, an \textit{unlearning algorithm} \(U(\cdot)\) consumes \(A(D)\) (the model trained on \(D\) using algorithm \(A\)) to facilitate the forgetting of set \(S\), thereby generating a model that has "forgotten" \(S\). The act of forgetting is assessed by the execution of the training algorithm on \(D\setminus S\) (i.e., \(A(D\setminus S)\), a model retrained from scratch without the forget set), thereby furnishing a mathematical framework for machine unlearning. \newline
\textbf{Definition of Machine Unlearning\cite{sekhari2021remember}:} \textit{Given a fixed dataset \(D\), forget set \(S \subseteq D\), and a learning algorithm \(A\), an unlearning algorithm \(U\) is considered to unlearn with respect to \((D,S,A)\) if, for all regions \(R\), the following conditions hold:
\begin{center}
    \(\Pr[A(D\setminus S) \in R] \leq e^\varepsilon \Pr[U(A(D), S, D) \in R] + \delta,\)
\end{center}
and
\begin{center}
    \(\Pr[U(A(D), S, D) \in R] \leq e^\varepsilon \Pr[A(D\setminus S) \in R] + \delta\)\newline
\end{center}}

In essence, the definition above states that as the values of \(\varepsilon\) and \(\delta\) tend towards zero, it becomes increasingly difficult for an observer to distinguish between the models \(A(D\setminus S)\) (a model retrained from scratch without the forget set) and \(U(A(D),S,D)\) (an unlearned model). This observation suggests that the distributions of the two models, \(A(D\setminus S)\) and \(U(A(D),S,D)\), are highly similar, if not practically indistinguishable. This implies that a good unlearning algorithm is as close as to the model that was trained without the forget set, highlighting that it behaves as though it was never trained on the data in the forget set. Additionally, the unlearning process should not significantly compromise the accuracy of the model or its ability to generalize well.\newline

\subsubsection{\textbf{Proposed Methodology}}
We propose a straightforward yet potent approach for unlearning, namely unlearning via fine-tuning.\newline
\textbf{Definition of Unlearning via Fine-tuning:} \textit{Let \(D\) denote the entire dataset, \(S \subseteq D\) represent the forget set, and \(A(D)\) denote the model trained on \(D\). Subsequently, the retained set (\(L\)) is defined as \(L = D-S\). The unlearning algorithm \(U\) is then characterized as:
\begin{center}
    \(U = A(D) \xrightarrow{fine-tuned-on} L\)
\end{center}}

Unlearning via fine-tuning initiates with the original pre-trained model, which serves as a precise representation of the entire dataset. Subsequently, the pre-trained model undergoes a process of fine-tuning, or optimization, specifically tailored to the retained set for a predetermined number of epochs. This approach offers several notable advantages:

\begin{itemize}
    \item Given that the original model accurately captures the characteristics of the entire dataset, the subsequent fine-tuned model is poised to maintain this fidelity, thereby ensuring a robust representation of the overall data distribution.
    \item By fine-tuning the model to the retained set, we effectively modify the model's functionality and representation of the data within the forget set. This adjustment substantially mitigates the risk of extracting sensitive information from the forget set through adversarial attacks.
    \item Leveraging fine-tuning as an extension of the pre-existing model construction facilitates a streamlined development process and ensures computational efficiency, thereby rendering it a practical and effective approach.
\end{itemize}

\textbf{Unlearning via Zero Shot Teacher Student Approach:} Chundawat, V. S. et. al. in their paper introduce a methodology centered around the concept of zero-shot unlearning through a process called Gated Knowledge Transfer (GKT). This method is particularly designed to handle scenarios where certain information needs to be unlearned by a model without the need for retraining from scratch, which can be computationally prohibitive.

The model architecture employed involves a teacher-student framework augmented by a generator. The teacher model, trained on the original dataset, serves as a baseline that retains knowledge. The student model is a randomly initialized model with the same architecture as the teacher model. The student aims to learn the information from the teacher model, excluding the knowledge of specific classes that need to be forgotten. The generator produces pseudo data points from a noise vector. These points are utilized to guide the student model towards mimicking the teacher's knowledge selectively. The generator $G(\mathbf{z}; \phi)$ produces pseudo data points from a noise vector $\mathbf{z} \in \mathcal{N}(0, 1)$. The generator is trained to maximize the Kullback-Leibler (KL) divergence between the output probabilities of the teacher and student models.

Kullback-Leibler (KL) divergence between output probabilities of the teacher and student model is
\begin{equation}
D_{KL} \left( T(x_p) \| S(x_p) \right) = \sum_i t_p^{(i)} \log \frac{t_p^{(i)}}{s_p^{(i)}}
\end{equation}

To ensure that the student focuses on the retained classes while ignoring the forgotten classes, they have introduced an attention loss.
\begin{equation}
L_{at} = \sum_{i \in \mathcal{N}} \left\| f\left( A_i^{(t)} \right) \left\| \frac{A_i^{(t)}}{\left\| A_i^{(t)} \right\|_2} \right\|_2 - f\left( A_i^{(s)} \right) \left\| \frac{A_i^{(s)}}{\left\| A_i^{(s)} \right\|_2} \right\|_2 \right\|_2
\end{equation}

A crucial design element called the band-pass filter is introduced. This filter ensures that information regarding the forget class(es) does not reach the student model from the teacher model, allowing the student to learn only the information of the retain class(es). To limit the generator from transmitting information about the forget class(es), they introduce a filter $F$ before the generator. This filter evaluates all generated pseudo samples and screens them before passing them to the student.

The filter criterion for each sample is defined as:
\begin{equation}
F(x_p) = \prod_{i \in \mathcal{C}_f} \left( t_p^{(i)} < \epsilon \right)
\end{equation}
\section{Results}
The Generalized Model Classifier, which has an accuracy of 54\%, the Fine-tuned Model Classifier, which has an accuracy of 58\%, and the Unlearned Model Classifier, which has an accuracy of 58\%, are contrasted in the table. With an accuracy of 58\%, both the Fine-tuned and Unlearned Model Classifiers perform better than the Generalized Model. This result is consistent with our premise that model performance will be improved by optimizing for personalized scenarios and using the unlearning procedure we are holding the accuracy along with that safeguarding the models from adversarial attacks. The actual findings support our theoretical predictions and show how these methods are useful for increasing accuracy.

\begin{table}[h]
\centering
\begin{tabular}{|l|c|c|}
\hline
\textbf{Models} & \textbf{Learning via Unlearning} & \textbf{Zero-shot Unlearning} \\ \hline
Generalized Model Classifier & 54\% & 62\% \\ 
Personalized Model Classifier & 58\% & 67.23\% \\ 
Unlearned Model Classifier & 58\% & 65.76\% \\ \hline
\end{tabular}
\caption{Comparison of Model Accuracies}
\label{table:model_accuracies}
\end{table}

The following figure shows the accuracy and loss curve for the generalized classifier model. The observed trend of decreasing loss and increasing accuracy with each epoch signifies a positive trajectory in the model's learning process. This pattern suggests effective learning and adaptation by the model to the training data, indicating the optimization process is successfully refining the model's parameters. These outcomes are indicative of a well-designed training regimen that aligns with best practices in machine learning, reinforcing the model's potential for achieving high performance on unseen data.

\begin{figure}[htbp]
    \centering
    \includegraphics[width=0.5\textwidth]{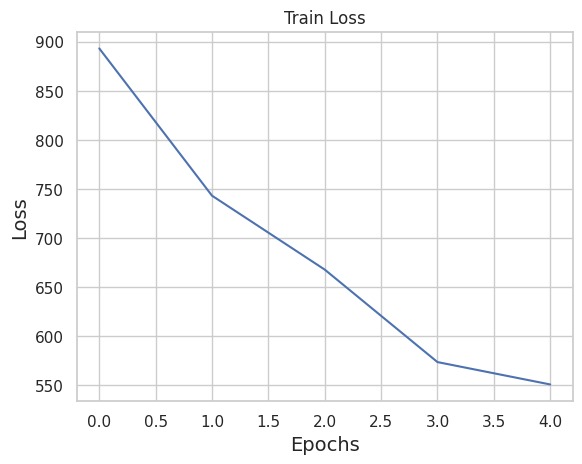}
    \caption{Loss Curve for Generalized Classifier Model }
    \label{Loss Curve for Generalized Classifier Model}
\end{figure}

\begin{figure}[htbp]
    \centering
    \includegraphics[width=0.5\textwidth]{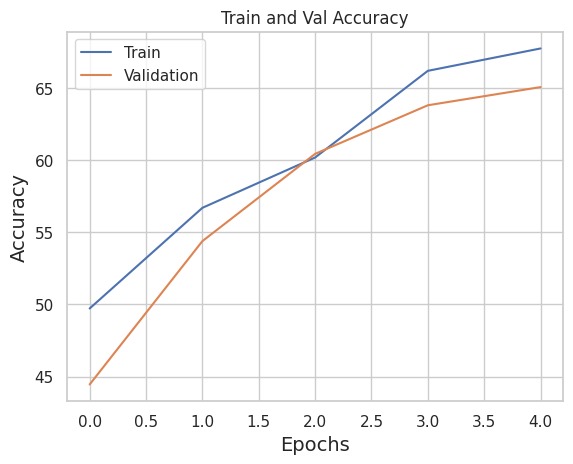}
    \caption{Accuracy Curve for Generalized Classifier Model }
    \label{Accuracy Curve for Generalized Classifier Model}
\end{figure}

The below results illustrate the losses incurred by:
\begin{enumerate}
    \item The Personalized Model vs. the Unlearned Model.
    \item The General Model vs. the Unlearned Model.
\end{enumerate}
To maintain consistency with the previously established convention, the Personalized Model is denoted as \(D\), here the forget set \(S\) comprises the patients' data, and the General Model is represented as \(D\setminus S\). \newline

\begin{figure}[htbp]
    \centering
    \includegraphics[width=0.5\textwidth]{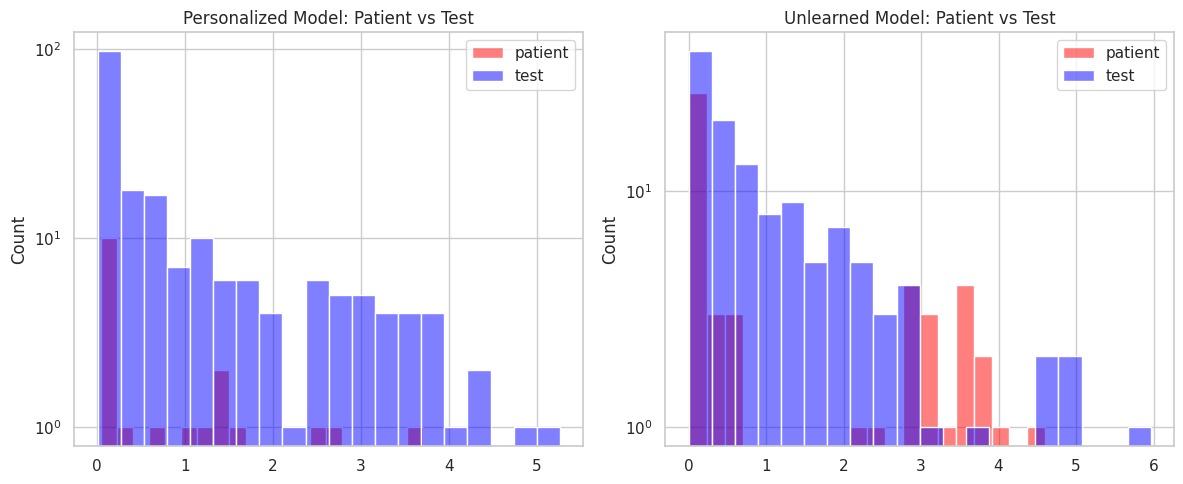}
    \caption{Test and Patient losses of Personalised Vs Unlearned model }
    \label{fig: Test and Patient losses of Personalised Vs Unlearned model }
\end{figure}
The Personalized Model undergoes fine-tuning on the patients' data, resulting in significantly lower losses compared to the test data. Conversely, the Unlearned Model is designed to remove the patients' data from its training, resulting in higher losses for patient data and a more dispersed distribution.\newline

\begin{figure}[htbp]
    \centering
    \includegraphics[width=0.5\textwidth]{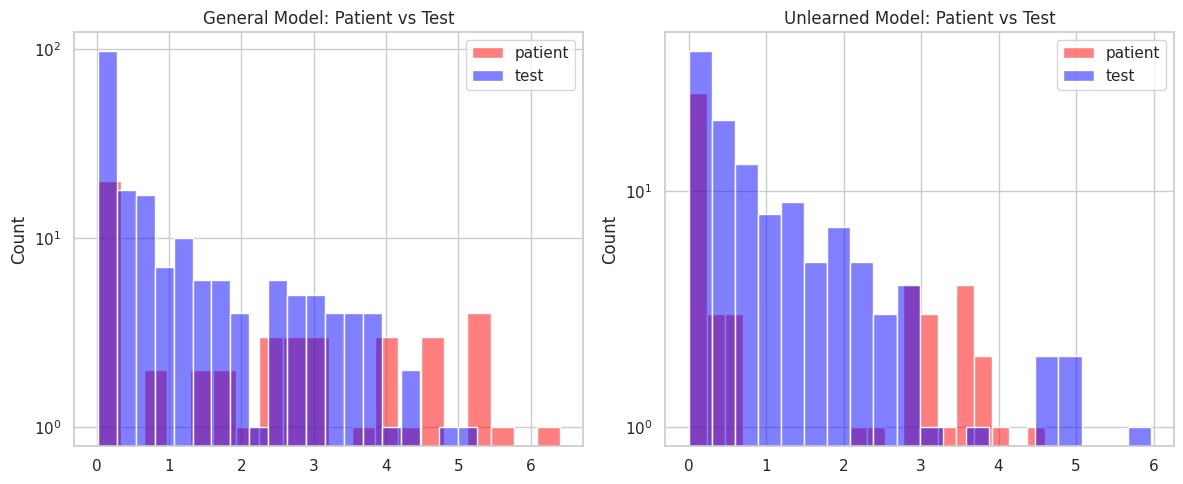}
    \caption{Test and Patient losses of General Vs Unlearned model}
    \label{fig: Test and Patient losses of General Vs Unlearned model}
\end{figure}
The General Model, trained solely on non-patient data, exhibits high and dispersed losses for patient data, consistent with its lack of exposure to this subset. Similarly, the Unlearned Model, resembling the General Model in its absence of patient data, demonstrates comparable outcomes, confirming the effectiveness of the unlearning algorithm.

\section{Conclusion}

In our empirical analysis, we observed a noteworthy similarity in the loss distribution between the Unlearned model and the Generalized model. This convergence in loss distributions suggests that the Unlearned model exhibits a behavior akin to "forgetting" the patient data, indicating a shift towards a more generalized representation. Despite this apparent similarity, it is crucial to highlight that the Unlearned model outperforms the Generalized model in terms of accuracy.\newline
This superiority in performance by the Unlearned model implies that, while it appears to have distanced itself from the specific characteristics of the patient data, it has managed to achieve a higher level of accuracy when compared to the Generalized model. The divergence in characteristics between patient-specific data and generalized data models has effectively mitigated the impact of adversarial attacks. This convergence towards the generalized model's output complicates the identification of new data points within personalized ECG models, enhancing data security and model robustness against such attacks. This intriguing result prompts further investigation into the underlying mechanisms at play, raising questions about the nature of the learned representations and their impact on model performance.


\bibliography{acml24}

\begin{thebibliography}{16}
\providecommand{\natexlab}[1]{#1}
\providecommand{\url}[1]{\texttt{#1}}
\expandafter\ifx\csname urlstyle\endcsname\relax
  \providecommand{\doi}[1]{doi: #1}\else
  \providecommand{\doi}{doi: \begingroup \urlstyle{rm}\Url}\fi

\bibitem[Baloglu et~al.(2019)Baloglu, Talo, Yildirim, San~Tan, and Acharya]{baloglu2019classification}
Ulas~Baran Baloglu, Muhammed Talo, Ozal Yildirim, Ru~San~Tan, and U~Rajendra Acharya.
\newblock Classification of myocardial infarction with multi-lead ecg signals and deep cnn.
\newblock \emph{Pattern recognition letters}, 122:\penalty0 23--30, 2019.

\bibitem[Bourtoule et~al.(2021)Bourtoule, Chandrasekaran, Choquette-Choo, Jia, Travers, Zhang, Lie, and Papernot]{bourtoule2021machine}
Lucas Bourtoule, Varun Chandrasekaran, Christopher~A Choquette-Choo, Hengrui Jia, Adelin Travers, Baiwu Zhang, David Lie, and Nicolas Papernot.
\newblock Machine unlearning.
\newblock In \emph{2021 IEEE Symposium on Security and Privacy (SP)}, pages 141--159. IEEE, 2021.

\bibitem[Cao and Yang(2015)]{cao2015towards}
Yinzhi Cao and Junfeng Yang.
\newblock Towards making systems forget with machine unlearning.
\newblock In \emph{2015 IEEE symposium on security and privacy}, pages 463--480. IEEE, 2015.

\bibitem[Carlini and Wagner(2017)]{carlini2017evaluating}
Nicholas Carlini and David Wagner.
\newblock Towards evaluating the robustness of neural networks, 2017.

\bibitem[Ginart et~al.(2019)Ginart, Guan, Valiant, and Zou]{ginart2019making}
Antonio Ginart, Melody Guan, Gregory Valiant, and James~Y Zou.
\newblock Making ai forget you: Data deletion in machine learning.
\newblock \emph{Advances in neural information processing systems}, 32, 2019.

\bibitem[Goodfellow et~al.(2015)Goodfellow, Shlens, and Szegedy]{goodfellow2015explaining}
Ian~J. Goodfellow, Jonathon Shlens, and Christian Szegedy.
\newblock Explaining and harnessing adversarial examples, 2015.

\bibitem[He et~al.(2016)He, Zhang, Ren, and Sun]{7780459}
Kaiming He, Xiangyu Zhang, Shaoqing Ren, and Jian Sun.
\newblock Deep residual learning for image recognition.
\newblock In \emph{2016 IEEE Conference on Computer Vision and Pattern Recognition (CVPR)}, pages 770--778, 2016.
\newblock \doi{10.1109/CVPR.2016.90}.

\bibitem[Jun et~al.(2018)Jun, Nguyen, Kang, Kim, Kim, and Kim]{jun2018ecg}
Tae~Joon Jun, Hoang~Minh Nguyen, Daeyoun Kang, Dohyeun Kim, Daeyoung Kim, and Young-Hak Kim.
\newblock Ecg arrhythmia classification using a 2-d convolutional neural network, 2018.

\bibitem[Kiranyaz et~al.(2015)Kiranyaz, Ince, and Gabbouj]{kiranyaz2015real}
Serkan Kiranyaz, Turker Ince, and Moncef Gabbouj.
\newblock Real-time patient-specific ecg classification by 1-d convolutional neural networks.
\newblock \emph{IEEE Transactions on Biomedical Engineering}, 63\penalty0 (3):\penalty0 664--675, 2015.

\bibitem[Kurakin et~al.(2017)Kurakin, Goodfellow, and Bengio]{kurakin2017adversarial}
Alexey Kurakin, Ian Goodfellow, and Samy Bengio.
\newblock Adversarial examples in the physical world, 2017.

\bibitem[Madry et~al.(2019)Madry, Makelov, Schmidt, Tsipras, and Vladu]{madry2019deep}
Aleksander Madry, Aleksandar Makelov, Ludwig Schmidt, Dimitris Tsipras, and Adrian Vladu.
\newblock Towards deep learning models resistant to adversarial attacks, 2019.

\bibitem[Sekhari et~al.(2021)Sekhari, Acharya, Kamath, and Suresh]{sekhari2021remember}
Ayush Sekhari, Jayadev Acharya, Gautam Kamath, and Ananda~Theertha Suresh.
\newblock Remember what you want to forget: Algorithms for machine unlearning, 2021.

\bibitem[Shi et~al.(2023)Shi, Ghalyan, Gourgoulias, Buford, and Moran]{shi2023deepclean}
Jiaeli Shi, Najah Ghalyan, Kostis Gourgoulias, John Buford, and Sean Moran.
\newblock Deepclean: Machine unlearning on the cheap by resetting privacy sensitive weights using the fisher diagonal.
\newblock \emph{arXiv preprint arXiv:2311.10448}, 2023.

\bibitem[Szegedy et~al.(2014)Szegedy, Zaremba, Sutskever, Bruna, Erhan, Goodfellow, and Fergus]{szegedy2014intriguing}
Christian Szegedy, Wojciech Zaremba, Ilya Sutskever, Joan Bruna, Dumitru Erhan, Ian Goodfellow, and Rob Fergus.
\newblock Intriguing properties of neural networks, 2014.

\bibitem[Wagner et~al.(2020)Wagner, Strodthoff, Bousseljot, Kreiseler, Lunze, Samek, and Schaeffter]{article}
Patrick Wagner, Nils Strodthoff, Ralf-Dieter Bousseljot, Dieter Kreiseler, Fatima Lunze, Wojciech Samek, and Tobias Schaeffter.
\newblock Ptb-xl, a large publicly available electrocardiography dataset.
\newblock \emph{Scientific Data}, 7:\penalty0 154, 05 2020.
\newblock \doi{10.1038/s41597-020-0495-6}.

\bibitem[Zhao et~al.(2020)Zhao, Cheng, Zhang, and Peng]{zhao2020ecg}
Yunxiang Zhao, Jinyong Cheng, Ping Zhang, and Xueping Peng.
\newblock Ecg classification using deep cnn improved by wavelet transform.
\newblock \emph{Computers, Materials and Continua}, 2020.

\end{thebibliography}

\end{document}